\documentclass[sigconf]{acmart}
\usepackage{enumitem}
\AtBeginDocument{%
  }

\copyrightyear{2025}
\acmYear{2025}
\setcopyright{cc}
\setcctype{by}
\acmConference[RecSys '25]{Proceedings of the Nineteenth ACM Conference on Recommender Systems}{September 22--26, 2025}{Prague, Czech Republic}
\acmBooktitle{Proceedings of the Nineteenth ACM Conference on Recommender Systems (RecSys '25), September 22--26, 2025, Prague, Czech Republic}\acmDOI{10.1145/3705328.3748015}
\acmISBN{979-8-4007-1364-4/2025/09}

\begin{document}

\title[Understanding LLM-generated Group Recommendations]{Consistent Explainers or Unreliable Narrators? Understanding LLM-generated Group Recommendations}

%the Understandability of Social Choice-based Explanations for Group Recommendation

%% I don't need a better explanation, 
% With friends like these, who needs explanations? 

\author{Cedric Waterschoot}
\affiliation{%
  \institution{Maastricht University}
  \city{Maastricht}
  \country{The Netherlands}}
\email{cedric.waterschoot@maastrichtuniversity.nl}

\author{Nava Tintarev}
\affiliation{%
  \institution{Maastricht University}
  \city{Maastricht}
  \country{The Netherlands}}
\email{n.tintarev@maastrichtuniversity.nl}

\author{Francesco Barile}
\affiliation{%
  \institution{Maastricht University}
  \city{Maastricht}
  \country{The Netherlands}}
\email{f.barile@maastrichtuniversity.nl}

\renewcommand{\shortauthors}{Waterschoot et al.}

\begin{abstract}

Large Language Models (LLMs) are increasingly being implemented as joint decision-makers and explanation generators for Group Recommender Systems (GRS). In this paper, we evaluate these recommendations and explanations by comparing them to social choice-based aggregation strategies. Our results indicate that LLM-generated recommendations often resembled those produced by \textit{Additive Utilitarian} (ADD) aggregation. However, the explanations typically referred to averaging ratings (resembling but not identical to ADD aggregation). Group structure, uniform or divergent, did not impact the recommendations. Furthermore, LLMs regularly claimed additional criteria such as user or item similarity, diversity, or used undefined popularity metrics or thresholds. Our findings have important implications for LLMs in the GRS pipeline as well as standard aggregation strategies. Additional criteria in explanations were dependent on the number of ratings in the group scenario, indicating potential inefficiency of standard aggregation methods at larger item set sizes. Additionally, inconsistent and ambiguous explanations undermine transparency and explainability, which are key motivations behind the use of LLMs for GRS. %Additionally, the number of individual ratings in a presented group scenario influences both the aggregation procedure as well as the explanation. %Finally, differences in recommendations between LLMs highlight how model choice influences outcomes. 

\end{abstract}




\maketitle

\section{Introduction}
Large Language Models (LLMs) are increasingly employed for recommendation, whether to generate explanations \cite{Lubos2024llmgen,Said2025explaining} or as decision-maker \cite{bao-etal-2024-decoding,Liao2024llara}. LLMs can be employed for their combined purpose, conflating the recommendation and explanation task. For the specific case of group recommender systems (GRS), LLMs have been employed to derive a group recommendation from a set of users, together forming a group, and their individual ratings of items \cite{Tommasel2024-dp,Waterschoot2025Pitfalls}. The growing use of LLMs requires clear evaluation of their performance as consistent decision-maker and explainer. 

However, group scenarios present a difficult challenge. The literaturedescribes aggregating individual preferences into a singular group recommendation as a complex task for which no universally suitable approach exists. \citet{barile2023evaluating} conclude that aggregation procedures, described under the umbrella of social choice-based aggregation strategies, are perceived differently in terms of fairness, consensus and satisfaction depending on the group.  

In the context of GRS, it remains unclear what LLM-generated recommendations look like and whether that derived output resembles group recommendations produced by applying a social choice-based aggregation strategy. Thus, a contextualization of LLM-generated group recommendations is needed to compare LLM output with aggregation strategies. Additionally, different LLMs need to be compared to investigate whether models follow similar procedures. Finally, we are interested in whether group-specific factors, such as (dis)similarity among group members and the number of potential items presented to the LLM, influence LLMs as decision-maker. We formulate the following research questions:\newline

\noindent \textbf{RQ1.} Do LLM-generated group recommendations match those derived from different social choice-based aggregation strategies?

\noindent \textbf{RQ2.} Does the group structure (uniform or divergent preferences) affect LLM performance?

\noindent \textbf{RQ3.} Do LLMs claim to have followed a specific aggregation procedure when prompting to generate explanations of the group recommendation?\newline

Our contributions can be summarized as follows. To the best of our knowledge, this study is the first to contextualize group recommendations generated by multiple LLMs across a range of social choice-based aggregation strategies, rather than relying on a single baseline. We extend our analysis by comparing LLM-generated recommendations for both uniform and divergent user groups. Lastly, we introduce a categorization of LLM-generated explanations of aggregation procedures, specifically investigating whether the claimed procedure is impacted by the group scenario presented to the LLM.

\section{Related Work}\label{sec:background}

%\subsection{Group Recommendation}\label{sec:GRS}
Group Recommender Systems (GRS) extend traditional recommender systems to process the preferences of multiple users, generating a single output suited for the group \cite{Masthoff2022group}. Such applications have previously been discussed in contexts such as music \cite{najafian2018generating}, restaurants \cite{barile2021toward} or tourism \cite{chen2021attentive}. To generate single recommendations rooted in a range of individual preferences, recent approaches employed methodologies such as attentive neural networks~\cite{cao2018attentive,huang2020efficient}, graph neural networks \cite{Zhang2021double} or reinforcement learning \cite{Stratigi2023squirrel}. 

An accessible procedure to derive a group recommendation from individual preferences are social choice-based aggregation strategies, rooted in \textit{Social Choice Theory} \cite{kelly2013social,masthoff2015group,masthoff2004group}. These strategies present distinct options to aggregate individual preferences into a group recommendation and have been widely employed as procedure \cite{barile2023evaluating,tran2019towards} or baseline of more complex approaches \cite{nguyen2019conflict,rossi2018altruistic,delic2018use}. Since these strategies offer diverging, explainable procedures to generate group recommendations, they are suitable as comparison to contextualize LLM-generated recommendations (e.g. as used by \cite{Tommasel2024-dp}). These strategies are typically categorized as \textit{consensus-based}, \textit{majority-based}, or \textit{borderline} \cite{senot2010analysis}. In this study, we use strategies from each category. The consensus-based strategy, \textit{Additive Utilitarian} (ADD), recommends the items with the highest sums of all ratings \cite{senot2010analysis}. The majority-based strategy, \textit{Approval Voting} (APP), selects the items with the most ratings above a set threshold \cite{senot2010analysis}. Finally, two borderline strategies are included: \textit{Least Misery} (LMS), which recommends the items with the highest of the lowest ratings, and \textit{Most Pleasure} (MPL), which selects the items with the overall highest rating \cite{senot2010analysis}.

\subsection{LLMs for (group) recommendation}
LLMs have been increasingly used for generating recommendations \cite{Zhao2024rec} to augment context awareness in the recommendation pipeline \cite{Xi2024openworld}, to generate explanations \cite{Lubos2024llmgen,Said2025explaining} or, similar to our case, as decision maker itself \cite{bao-etal-2024-decoding,Liao2024llara}. Prior work has utilized the interactive capabilities of LLMs to address cold-start problems \cite{Sanner2023-kb,Wu2024could} and to create conversational recommender systems \cite{Gao2023chatrec,Yang2024behav}. The possibilities resulting from LLMs and zero- and few-shot learning resulted in applications geared towards data sparse tasks including cross-domain recommendation \cite{Petruzzelli2024instruct,Kim2024large}. Overall, LLMs have shown promise and challenged conventional recommendation methodologies \cite{Hou2024large,dong2024survey}. Despite an increase in LLM-generated recommendations, it is unclear how LLMs derive recommendations when presented with group scenarios containing diverging preferences. Additionally, it remains to be seen whether different LLM models provide different recommendations or follow dissimilar aggregation procedures. Finally, we aim to investigate whether LLM-generated recommendations reflect those derived by social choice-based aggregation strategies, used as ground truth for evaluation of group recommendations \cite{Tommasel2024-dp,Waterschoot2025Pitfalls}.

%wever, widespread use of LLMs, especially on external servers or making use of APIs, has raised privacy concerns, due to the prominence of user-related data in recommendation scenarios \cite{Wu2023enhanced,Zhao2024rec}.  An alternative presents itself in the use of smaller or open-source models that can be ran locally, omitting the need for external computation or API use (see e.g. \cite{Wiest2024privacy}).

%The performance of LLM-based recommender systems for group recommendation remains uncertain. Previously, group recommendations generated by LLMs have been evaluated in terms of fairness and inclusion of sensitive attributes \cite{Tommasel2024-dp}. The authors evaluate movie recommendations made by three language models, with and without sensitive user attributes in the prompt, and compare it to a baseline by additive aggregation. However, although social choice-based aggregation strategies are commonly used to generate group recommendations, it is uncertain whether LLMs can effectively implement them. Additionally, the influence of scenario complexity -- the total number of ratings to process -- has not been systematically examined in the evaluation pipeline. This is particularly important since group information is embedded within the prompt alongside other instructions and risk being forgotten, leading to an incorrect output.

\section{Methodology}\label{sec:methods}
In the following paragraphs, we describe the methodology used to generate and analyze LLM-generated group recommendations and explanations. First, we outline how we created fictitious group scenarios. Second, we detail the overall LLM pipeline. Lastly, the procedure for answering each research question is described: (i) matching LLM output with social choice-based aggregation strategies, (ii) calculating within-group distance to label uniform and divergent groups, and (iii) categorizing explanations.

\subsection{Group generation}\label{sec:groupgen}
For this particular experiment, we created a set of randomized, fictitious group scenarios. Such a scenario consists of a matrix $M$ with size $U \times I$, where $U$ is the number of group members and $I$ the number of items.
Inspired by previous work by \citet{barile2023evaluating} and \citet{Waterschoot2025withfriends}, we generated random ratings for $4$ members ranging between 0 and 100. We opted for a scale ranging between 0 and 100 (as opposed to 10-point scale) to decrease ties for scenarios including a high number of items. To account for the number of ratings LLMs ought to process, we included group scenarios containing either 25, 50 or 75 items, thus containing either 100, 200 or 300 individual user-item ratings. For each matrix size, we generated $500$ scenarios, resulting in $1,500$ distinct group scenarios. We anonymized both users ($User_x$ with $x \in [1,4]$) and items ($Item_x$ with $x \in [1,num\_items]$) in the scenarios. We did not include any domain cues such as restaurants or movie titles. The full code is found in the companion Github repository.\footnote{\label{fn:github}\url{https://github.com/Cwaterschoot/GRS-LLM-expl-unreliable}}

\subsection{LLM Pipeline}\label{sec:pipeline}

\subsubsection{Random baseline}
As a baseline, we included a random group recommendation. For every group, we generated a randomized top 10 list based on the items present in each distinct scenario.

\subsubsection{Prompt Construction}
We opted for a simple prompt construction. First, we introduced the goal: \textit{``You are an expert in making and explaining group recommendations based on the knowledge base provided below.''} Second, the prompt described the format of the group scenario: \textit{``The information includes users (user\_id) and information on items they like (item\_x). The rating is a scale from 0 to 100. When referring to items, use item\_value.''}. Afterwards, for each iteration, the group scenario was inserted between tags to separate prompt from group table, similar to previous work \cite{Waterschoot2025Pitfalls}. Finally, the prompt included output formatting instructions. LLMs were instructed to only return a JSON object containing the `recommendation' (top 10 item list) and `explanation' keys. The full prompt is found in the companion repository.\footnotemark[1]

\subsubsection{LLMs}

%To simplify our pipeline and ensure reproducibility, 
We used Ollama to run LLMs and used the \textit{Langchain} python package\footnote{Langchain version 0.3.2} to load and prompt the included models. In this study, we present recommendations generated by four LLMs: (1) Llama3.1 (\textit{llama3.1:8b-instruct-q8\_0}, 8B), (2) Mistral NeMo (\textit{mistral-nemo:12b\-instruct-2407-q4\_K\_M}, 12B), (3) Gemma3 (\textit{gemma3:12b}, 12B) and (4) Phi4 (14B). 
%These models vary in parameter size and can be ran locally using quantized versions. Local models omit the need for APIs or external servers and thus, present a valuable option to address privacy concerns due to the prominance of user-related information in recommendation scenarios~\cite{Wu2023enhanced,Zhao2024rec,Wiest2024privacy}.

\subsection{RQ1: LLM Aggregation Strategies}
%To contextualize and evaluate the LLM-generated top 10 group recommendations, we made use of four different social choice-based aggregation strategies as ground truth. This allowed us to investigate whether the LLM output was similar to the often used strategies to derive group recommendations. 
To answer RQ1, we compared the generated recommendations to ground truth based on the \textit{Additive Utilitarian} (ADD), \textit{Most Pleasure} (MPL), \textit{Least Misery} (LMS) and \textit{Approval Voting} (APP) strategies~\cite{senot2010analysis,masthoff2015group,barile2023evaluating}. We calculated NDCG@10 between the top 10 recommendations generated by each LLM and each strategy.

\subsection{RQ2: Effect of Group Structure}
%To address RQ2, we were interested in investigating potential differences in LLM performance when presented with similar (group pattern where users preferences align) or dissimilar (higher distance between user preferences) groups. 
To answer RQ2, we categorized the randomly generated user groups (as introduced in Section \ref{sec:groupgen}) using a method adapted from previous work \cite{barile2023evaluating,Waterschoot2025withfriends}. While this earlier work used distance or correlation \cite{Kaya2020-fs}) to select users while generating groups, we calculate within-group distance of already generated groups. For each group $g$, we first calculated a distance matrix $D_{g}$, each entry representing the Euclidean distance between two users. Two users with similar preferences resulted in a smaller distance. To obtain the overall similarity within a group, we averaged and normalized the values in the distance matrix $D_{g}$, resulting in the group's internal user distance. To classify a group as uniform or divergent, we used a threshold based on the normal distribution of the within-group distances. We categorized group $g$ as \textit{uniform} if its distance was below $\mu_{dist_{g}} - \sigma_{dist_{g}}$, and as \textit{divergent} if it was above threshold $\mu_{dist_{g}} + \sigma_{dist_{g}}$, where $\mu$ and $\sigma$ denote the mean and standard deviation of the normalized within-group distances. Groups falling between both thresholds were discarded. This resulted in $242$ uniform and $236$ divergent groups. \footnote{Detailed statistics and visualizations of distributions and thresholds are found in the companion repository: \url{https://github.com/Cwaterschoot/GRS-LLM-expl-unreliable}} To evaluate LLM performance based on group composition, we calculated the difference in average NDCG@10 scores between divergent and uniform groups ($\Delta NDCG@10$). This comparison was performed separately for each LLM-strategy pair (e.g., Llama-ADD or Phi-APP). 

%using the Euclidean distance between the ratings vector $R_{gu}$ for each user $u$ in group $g$. The upper triangle of each distance Matrix $D_{g}$ was averaged and normalized to obtain the within-group user distance $d_{norm_g}$. We categorized groups using threshold $\mu_{d_{norm_g}} \pm \sigma_{d_{norm_g}}$ where $\mu_{d_{norm_g}}$ is the mean normalized within-group distances and $\sigma_{d_{norm_g}}$ equal to the standard deviation. Each group $g$ for which $d_{norm_g}$ was above the upper distance threshold was categorized as \textit{dissimilar}, groups under the lower distance threshold were categorized as \textit{similar}. Groups for which we calculated a $d_{norm_g}$ between the lower and upper distance threshold received the \textit{neutral} label. We ended up with $242$ similar, $236$ dissimilar and $1,022$ neutral groups. A more detailed outline of the distances and visualizations of both distributions and thresholds are found in the companion repository \footnote{URL OF ANON REPO HERE.}. To compare performance on similar and dissimilar groups, we calculated $\Delta NDCG@10$ by subtracting the average NDCG@10 for dissimilar groups from the average NDCG@10 score calculated for similar groups. We did this calculation separately for every LLM-strategy (e.g. Phi-ADD, Phi-APP) pair.

\begin{table*}[htb]
    \small
  \caption{Average NDCG@10 scores of LLMs calculated with social choice-based aggregation strategies (ADD = Additive Utilitarian, MPL = Most Pleasure, LMS = Least Misery, APP = Approval Voting) as baseline; either 25, 50 or 75 items. Included LLMs are Llama, Mistral, Gemma and Phi and a randomized recommendation.}
  \label{tab:overallndcg}
  \centering
  \begin{tabular}{ccccccccccccc}
    \toprule
    & \multicolumn{3}{c}{ADD} & \multicolumn{3}{c}{MPL} & \multicolumn{3}{c}{LMS} & \multicolumn{3}{c}{APP}   \\
    \cmidrule(lr){2-4} \cmidrule(lr){5-7} \cmidrule(lr){8-10} \cmidrule(lr){11-13}
    & 25 & 50 & 75 & 25 & 50 & 75 & 25 & 50 & 75 & 25 & 50 & 75\\
    \midrule
    Llama & \textbf{0.92}&0.81 & 0.77& 0.88&0.73 &0.69 &\textbf{0.82} &\textbf{0.72} &0.67 &0.86 & 0.74&0.69 \\
    Mistral & 0.89&0.78 &0.74 & 0.86& 0.73&0.66 &\textbf{0.82} &\textbf{0.72} & 0.65&0.84 &0.73 &0.69 \\
    Gemma & 0.91&0.82 &0.72 &\textbf{0.95} &\textbf{0.84} & 0.72&0.77 &0.66 & 0.58& \textbf{0.88}& 0.76&0.64 \\
    Phi & 0.91&\textbf{0.84} & \textbf{0.81}&0.91 & 0.81& \textbf{0.78}&0.79 &0.71 & \textbf{0.69}& 0.86& \textbf{0.78}& \textbf{0.72}\\
    \midrule
    Random & 0.67& 0.52&0.41 &0.66 &0.52 &0.42 &0.67 &0.52 & 0.43& 0.67&0.50 & 0.40\\
    \bottomrule
  \end{tabular}
\end{table*}

\subsection{RQ3: Explanation Categorization}\label{sec:rulebasedcat}
Our prompt contained additional instructions to generate natural language explanations describing how the group recommendation was derived. More specifically, we prompted for \textit{``an explanation and example of your recommendation procedure, which someone with no knowledge of recommender systems could understand''}. We categorized these LLM-generated explanations according to fixed labels and rules. Given the formulaic nature of these texts, we used a rule-based categorization approach, matching the explanations to categories describing procedures, instead of more complex methods like embedding similarity. Fuzzy string matching, previously used in similar classification tasks \cite{Pikies2021analysis,Kumar2019fuzzy,hosseini-etal-2020-deezymatch}, allowed for controlled, interpretable classification, as key terms were predictable. Keywords were based on social choice-based explanations (e.g.,\cite{najafian2018generating,kapcak2018tourexplain, felfernig2018explanations}) and basic aggregation terms such as ``\textit{average}'' or ``\textit{user similarity}''.\footnote{Full keyword lists in repository.}

We used the RapidFuzz library to apply fuzzy string matching with Levenshtein distance (similarity threshold = $0.85$) \cite{pettersson-etal-2013-normalisation} and allowed for multiple labels per explanation. To improve accuracy, we added a negation check (e.g., to capture \textit{``did not average''} and number extraction (e.g., to equate thresholds like \textit{``ratings above [value]''}. We validated our rule-based approach by manually annotating 100 randomly selected explanations across LLMs and item sizes. This resulted in a Cohen's $\kappa = 0.76$, indicating substantial agreement with the assigned labels.

\section{Analysis and Discussion}\label{sec:results}
In the following paragraphs, we first report the overall resemblance of LLMs compared to social choice-based aggregation strategies (RQ1). Afterwards, we outline performance on uniform and divergent groups (RQ2) and discuss the categorization of the generated explanations (RQ3) .

\subsection{RQ1: Overall Resemblance}\label{sec:overallperformance}
Table \ref{tab:overallndcg} summarizes the average NDCG@10 scores comparing LLM-generated recommendations with those derived by applying social choice-based aggregation strategies. Generally speaking, LLMs outperformed the random baseline across all conditions, indicating that LLMs perform a non-random recommendation procedure (Table \ref{tab:overallndcg}). However, our results show important distinctions between different LLMs regarding their performance compared to social choice-based aggregation strategies.

In contrast to other LLMs, the recommendations produced by \textit{Gemma3} resulted in a higher NDCG@10 score for the Most Pleasure (MPL) strategy compared to the other included social choice-based strategies (Table \ref{tab:overallndcg}). The three other LLMs were consistently resulting in recommendations similar to the Additive Utilitarian (ADD). 

Unsurprisingly, performance dropped when the number of items increased due to rising scenario complexity and increasing options. Additionally, when accounting for 75 items, \textit{Phi4} outperformed all other LLMs for each of the four strategies. Interestingly, this trend was not found at lower item numbers. When presented with 50 items, \textit{Phi4} achieved the optimum NDCG@10 score for the ADD and APP strategies, while at 25 items it did not reach the optimum NDCG@10 score for any of the included strategies. This result showcases the importance of including a varying degree of item counts, as it can impact the relative performance across LLMs.

Thus, evaluation based on a singular aggregation strategy might create an incorrect picture of the recommendation capabilities of different LLMs. We have seen in earlier work that certain strategies are perceived as more fair or satisfactory depending on the specific group structure \cite{barile2023evaluating}. If an LLM-generated recommendation does not align with one strategy but overlaps with another, it may still be correct output for a given group scenario. Evaluating LLM-generated group recommendations using multiple aggregation procedures captures this outcome.

\begin{comment}
\begin{table*}[h]
    \small
  \caption{Mean Average Precision@10 (MAP@10) scores of LLMs compared to social choice-based aggregation strategies as ground truth (ADD = Additive Utilitarian, MPL = Most Pleasure, LMS = Least Misery, APP = Approval Voting) and random recommendation; either 25, 50 or 75 items. Included LLMs are Llama, Mistral, Gemma and Phi.}
  \label{tab:overallmap}
  \centering
  \begin{tabular}{ccccccccccccc}
    \toprule
    & \multicolumn{3}{c}{ADD} & \multicolumn{3}{c}{MPL} & \multicolumn{3}{c}{LMS} & \multicolumn{3}{c}{APP}   \\
    \cmidrule(lr){2-4} \cmidrule(lr){5-7} \cmidrule(lr){8-10} \cmidrule(lr){11-13}
    & 25 & 50 & 75 & 25 & 50 & 75 & 25 & 50 & 75 & 25 & 50 & 75\\
    \midrule
    Llama & 0.57&0.35 &0.31 & 0.47&0.23 & 0.19&\textbf{0.41} &\textbf{0.25} &0.20 &0.47 & 0.27&0.22 \\
    Mistral & 0.48&0.31 & 0.18& 0.42& 0.23& 0.13&0.39 &0.23& 0.13&0.42 &0.24 & 0.15\\
    Gemma & 0.55&0.36 &0.18 &\textbf{0.60} &\textbf{0.36} & 0.18&0.34 &0.18 &0.11 & \textbf{0.51}& 0.28& 0.14\\
    Phi & \textbf{0.61}&\textbf{0.42} & \textbf{0.37}&0.54 & 0.33&\textbf{0.27} &0.39 &\textbf{0.25} & \textbf{0.21}& \textbf{0.51}& \textbf{0.33}& \textbf{0.26}\\
    \midrule
    Random & 0.22& 0.08& 0.05&0.22 &0.09 &0.05 &0.22 &0.08 &0.05 & 0.22&0.08 &0.05 \\
    \bottomrule
  \end{tabular}
\end{table*}
\end{comment}

\subsection{RQ2: Uniform versus Divergent Groups}

\begin{figure*}[h]
  \centering
  \includegraphics[width=1\linewidth]{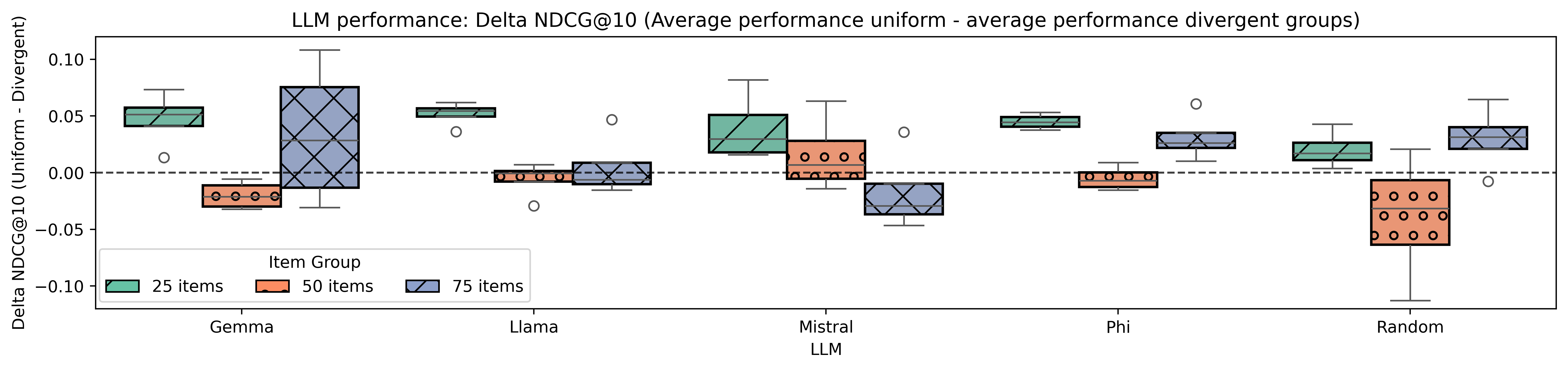}
  \caption{Average $\Delta NDCG@10$ scores for uniform and divergent groups (four values per LLM, one for each strategy). Split up by number of items in the group scenarios (25, 50 or 75 items). A positive $\Delta NDCG@10$ indicates better performance on uniform groups compared to divergent groups. 0 means equal performance.}
  \label{fig:delta-ndcg}
  \Description{Average differences between NDCG@10 scores for uniform and divergent groups. We did not find clear patterns and differences between uniform and divergent groups, as indicated by average differences in NDCG@10 scores.}
\end{figure*}

Figure \ref{fig:delta-ndcg} summarizes the difference in LLM performance between uniform and divergent groups. Defined as $\Delta NDCG@10$, this evaluation was obtained by subtracting the average NDCG@10 scores for each model-strategy pair for divergent groups with the value obtained for uniform groups. As a result, each LLM has a distinct $\Delta NDCG@10$ for the ADD, MPL, LMS and APP strategies, visualized in Figure \ref{fig:delta-ndcg}. A positive $\Delta NDCG@10$ indicates better performance on uniform groups, zero indicating equal performance.

Generally speaking, LLMs performed slightly better on uniform groups compared to divergent ones. This trend was found across the board for 25-item groups. Results for 50 and 75-item groups were more varied. Overall, the differences remained fairly limited (Figure \ref{fig:delta-ndcg}). However, the outliers do illustrate the value of including differing group configurations in GRS analysis. For example, we calculated a $\Delta NDCG@10$ for the \textit{Gemma3-LMS} pair on 75-item groups of $0.11$. Out of the 48 $\Delta NDCG@10$ values (4 LLMs x 4 Strategies x 3 Item counts), 12 exceeded $0.05$ (five belonging to \textit{Gemma3}, three to \textit{Llama3.1} and two derived from both \textit{Mistral} and \textit{Phi4}).

\subsection{RQ3: Explanation Categorization}\label{sec:explanation-results}
%\begin{comment}
    
\begin{table*}[htb]
    %\small
  \caption{Percentage of explanations categorized by a selection of labels: averaging ratings (Avg.), generating diversity (Div.), using similarity between users (Sim.), and using an undefined popularity threshold (Undef. pop). Results are  reported by LLM (Llama, Mistral, Gemma, and Phi) and the number of items in the scenario (25, 50 or 75 items).}
  \label{tab:categories}
  \centering
  \begin{tabular}{ccccccccccccc}
    \toprule
    & \multicolumn{3}{c}{Avg.} & \multicolumn{3}{c}{Sim.} & \multicolumn{3}{c}{Div.} & \multicolumn{3}{c}{Undef. pop.}   \\
    \cmidrule(lr){2-4} \cmidrule(lr){5-7} \cmidrule(lr){8-10} \cmidrule(lr){11-13}
    &  25 & 50 & 75 & 25 & 50 & 75 & 25 & 50 & 75 & 25 & 50 & 75\\
    \midrule
    Llama & 21.4&40.6 &73.8 &26.2 &43.2 &74.8 &14.8 &13.0 &5.6 &76.2 &55.8 &23.0 \\
    Mistral & 67.4& 86.4 &78.4 & 1.4&4.8 &0.8 &11.2 &35.8 &63.8 &24.6 &10.2 &21.6 \\
    Gemma & 74.2&83.6 &79.2 &4.8 &0.6 &1.2 &26.8 &23.2 &26.0 & 23.8&15.8 &18.0 \\
    Phi & 97.2& 98.6& 94.4& 0.2& 0.2& 0.8& 28.0& 30.4& 46.2& 2.2&1.4 &6.4 \\

    \bottomrule
  \end{tabular}
\end{table*}
%\end{comment}

\begin{comment}

\begin{table}[htb]
    %\small
  \caption{Percentage of explanations categorized by a selection of labels (Averaging ratings (Avg.), using similarity between users (Sim.), generating diversity (Div.), and using an undefined popularity threshold (Undef. pop). Results are reported per LLM (Llama, Mistral, Gemma, and Phi)}
  \label{tab:categories}
  \centering
  \begin{tabular}{ccccc}
    \toprule
    & Avg. & Sim. & Div. & Undef. pop.   \\
    \midrule
    Llama & 45.27 &48.07 &11.13&51.67 \\
    Mistral & 77.40 & 2.33 &36.93 &18.80 \\
    Gemma & 79.00 &2.20 &25.33 & 19.20 \\
    Phi & 96.73& 0.40& 34.87& 3.33 \\

    \bottomrule
  \end{tabular}
\end{table}

\end{comment}

\begin{table}[htb]
    %\small
  \caption{Percentage of explanations categorized by a selection of labels (Averaging ratings (Avg.), similarity (Sim.), diversity (Div.), and undefined popularity threshold (Undef. pop). Results are reported per LLM (Llama, Mistral, Gemma, and Phi) and per group configuration (Divergent, Uniform)}
  \label{tab:categories-configuration}
  \centering
  \begin{tabular}{cccccc}
    \toprule
    & &Avg. & Sim. & Div. & Undef. pop.   \\
    \midrule
    Llama & Divergent&42.15 &48.34 &10.74&55.78 \\
    & Uniform&39.83 &45.34&7.63&55.51 \\
    Mistral &Divergent &75.63 & 3.31 &39.26 &19.83 \\
    & Uniform&75.42 &2.54&42.80&20.76 \\
    Gemma & Divergent&80.16 &1.24 &24.38 & 19.83 \\
    & Uniform&76.69 &2.12 &25.00&19.49 \\
    Phi &  Divergent&93.39& 0.49& 35.71& 2.64 \\
    & Uniform&96.61 &0.42 &33.47&3.34 \\

    \bottomrule
  \end{tabular}
\end{table}

Table \ref{tab:categories} summarizes the common category labels derived from the rule-based classification presented in Section \ref{sec:rulebasedcat}. The most common strategy was to average ratings per item and recommend to top 10 items with the highest average, especially apparent for \textit{Phi}. The prevalence of averaging ratings mirror the performance on the ADD strategy. Due to each item being rated by all four users, simply summing or averaging ratings per item leads to the same final recommendations (even if the ratings are different). However, with the exception of \textit{Phi}, we did find a lot of explanations categorized under ``Undefined popularity'', thus lacking any clear insight into how items were concluded to be popular. Examples of ``Undefined popularity'' are \textit{``I recommended the most popular items''} or \textit{``The recommendation includes items that are well-liked by the group''}. Such explanations do not clearly indicate how popular was defined or which threshold was used. We did not find clear differences or trends between the generated explanations in cases of uniform and divergent group scenarios (Table \ref{tab:categories-configuration}).\footnote{Exploratory analysis did not reveal statistically significant differences between the group configurations and LLMs}

Interestingly, as the number of items increased, the specificity of explanations generated by \textit{Llama} improved. We see this in a tendency towards more averaging labels, and fewer undefined popularity calculations (Table \ref{tab:categories}). Finally, as the number of presented items grew, \textit{Llama} started referencing user or item similarity more often in generated explanations, thus deviating from strictly averaging ratings. Additionally, an identical trend for claims of diversity was found for \textit{Mistral} and \textit{Phi} (Table \ref{tab:categories}). 

These findings could partly explain the decreasing performance compared to the ADD strategy as the number of items increased. LLMs were combining multiple procedures simultaneously, balancing the average scores in terms of similarity and diversity in the generated top 10 group recommendations.

\section{Impact and Conclusion}
LLMs are increasingly being used in the context of GRS. However, it remained unclear how LLMs aggregate individual preferences into a singular recommendation. In this paper, we contextualized LLM-generated group recommendations by comparing them to social choice-based aggregation strategies. Additionally, we compared LLM performance on uniform and divergent groups and labeled LLM-generated explanations based on the described procedure. 

Our findings have important implications for the implementation of LLMs in GRS. All in all, LLM-generated recommendations tended to resemble ADD aggregation, while LLM-generated explanations claimed to average ratings. However, we found important differences across LLMs both in resemblance to commonly used aggregation procedures and explanations. For example, explanations generated by \textit{Llama} regularly mentioned user similarity, while those generated by \textit{Mistral} and \textit{Phi} often explicitly stated diversity in the recommendation list. These results indicate that, opposed to applying social choice-based strategies, LLMs tend to combine multiple approaches to derive a group recommendation. While the output might not resemble that derived by applying a singular aggregation strategy, it might still be accepted by the group. 

Another implication of our work is the impact of the number of ratings on both recommendations and explanations. Unsurprisingly, NDCG@10 scores decreased across the board when the number of items increased, due to more complex group scenarios. Interestingly, the increasing mentions of similarity and diversity followed the increase in item set size. In terms of informativeness of explanations, the presence of ``\textit{Undefined popularity}'' decreased when the item set size increased. This instability needs to be acknowledged when using LLMs for GRS.

For this specific study, we generated fictional group data. This procedure allowed to compare uniform and divergent groups, as well as analyze LLM-generated group recommendations in scenarios with varying item set sizes. However, future work should include experimenting on real-world group data.

In short, LLMs are not consistent when deriving group recommendations from a set of individual ratings. We outlined different approaches among LLMs, and found an impact of item set size on both resemblance to social choice-based aggregation strategies and procedures described in LLM-generated explanations. 

%LLMs did not uniformly follow the same procedure compared to each other and were not consistent in the aggregation procedure. Our result show the importance of including multiple LLMs, different group structures and a varying degree of items presented in the prompt. Additionally, LLM-generated explanations might not be informative, illustrated by the presence of ``\textit{Undefined popularity}'' and vague descriptors such as diversity and similarity, which can not be verified directly. 

%\section{Conclusion}

%\section{Appendices}

\bibliographystyle{ACM-Reference-Format}
\bibliography{reprobib}

\appendix

\end{document}